\documentclass[10pt,twocolumn,letterpaper]{article}

\usepackage{iccv}
\usepackage{times}
\usepackage{epsfig}
\usepackage{graphicx}
\usepackage{amsmath}
\usepackage{amssymb}
\usepackage{multirow}
\usepackage{array}

\usepackage[T1]{fontenc}

\usepackage[pagebackref=true,breaklinks=true,letterpaper=true,colorlinks,bookmarks=false]{hyperref}

\iccvfinalcopy 

\def\iccvPaperID{2347} 

\ificcvfinal\fi
\begin{document}

\title{The iMaterialist Fashion Attribute Dataset}

\author{Sheng Guo$^{1}$\hspace{15pt}Weilin Huang$^{1}$\hspace{15pt}Xiao Zhang$^{2}$\hspace{15pt}Prasanna Srikhanta$^{3}$\hspace{15pt}Yin Cui$^{4}$\hspace{15pt}Yuan Li$^{5}$\\Matthew R. Scott$^{1} $\hspace{15pt}Hartwig Adam$^{2}$\hspace{15pt}Serge Belongie$^{4}$\\
$^{1}$Malong Technologies\hspace{20pt}$^{2}$Google AI\hspace{20pt}$^{3}$Wish\hspace{20pt}$^{4}$Cornell University\hspace{20pt}$^{5}$Horizon Robotics}

\maketitle

\begin{abstract}
Large-scale image databases such as ImageNet have significantly advanced image classification and other visual recognition tasks. However much of these datasets are constructed only for single-label and coarse object-level classification. For real-world applications, multiple labels and fine-grained categories are often needed, yet very few such datasets exist publicly, especially those of large-scale and high quality. In this work, we contribute to the community a new dataset called iMaterialist Fashion Attribute (iFashion-Attribute) to address this problem in the fashion domain. The dataset is constructed from over one million fashion images with a label space that includes 8 groups of 228 fine-grained attributes in total. Each image is annotated by experts with multiple, high-quality fashion attributes. The result is the first known million-scale multi-label and fine-grained image dataset. We conduct extensive experiments and provide baseline results with modern deep Convolutional Neural Networks (CNNs). Additionally, we demonstrate models pre-trained on iFashion-Attribute achieve superior transfer learning performance on fashion related tasks compared with pre-training from ImageNet or other fashion datasets. Data is available at: \url{https://github.com/visipedia/imat_fashion_comp}
\end{abstract}

\section{Introduction}
Recent deep learning models trained on large-scale datasets (\eg, ImageNet~\cite{DengDSLL009} and Open Images~\cite{kuznetsova2018open}) have significantly advanced the task of image classification and related applications in computer vision, such as object detection~\cite{ren2015faster,liu2016ssd,redmon2016you,lin2018focal}, segmentation~\cite{long2015fully,hong2015decoupled,he2017mask} and retrieval~\cite{hadi2015buy}. Performance on existing image classification benchmarks such as ImageNet~\cite{DengDSLL009} has reached the saturation point when using recent CNNs~\cite{szegedy2015going,HeK2016,szegedy2016rethinking,hu2018senet}. 
New datasets need to be created to tackle more challenging problems, such as multi-label classification and fine-grained recognition. 
On the other hand, domain-specific datasets have raised a lot of interest recently, especially in the fashion domain~\cite{xiao2015learning,liu2016deepfashion,hadi2015buy,yamaguchi2012parsing}. In light of this, we introduce the iMaterialist Fashion Attribute Dataset (iFashion-Attribute).
%
The iFashion-Attribute dataset includes over one million high-quality annotated fashion images. The label space includes 8 groups and a total of 228 fashion attributes. Details of the fashion groups and attribute-level classes are described in Table \ref{tbl:iMaterialist Fashion}. The labels are curated by fashion experts and each image has on average 5 individual labels.

iFashion-Attribute presents a few unique challenges. Firstly, it is a multi-label prediction problem and the models are evaluated by precision and recall. Multi-label image recognition has been studied in the community with recent deep learning based approaches~\cite{wei2016hcp,wang2016cnn,wang2017multi,li2017improving,zhu2017learning,lin2014microsoft}. However, performance of this task is significantly lower than that of ImageNet classification. Most existing datasets created for multi-label image recognition are limited in scale, such as PASCAL VOC~\cite{everingham2015pascal}, MS-COCO~\cite{lin2014microsoft} and NUS-WIDE~\cite{nuswide}, which have about 6K, 80K and 160K training images, with 20, 80 and 81 categories, respectively. Both learning difficulty and annotation effort would be increased considerably when the number of categories increases. It is particularly challenging to collect a large-scale (\eg, million-level) database to benchmark this task. 


\begin{figure*}[tb]
\centering
\includegraphics[width=1.6\columnwidth]{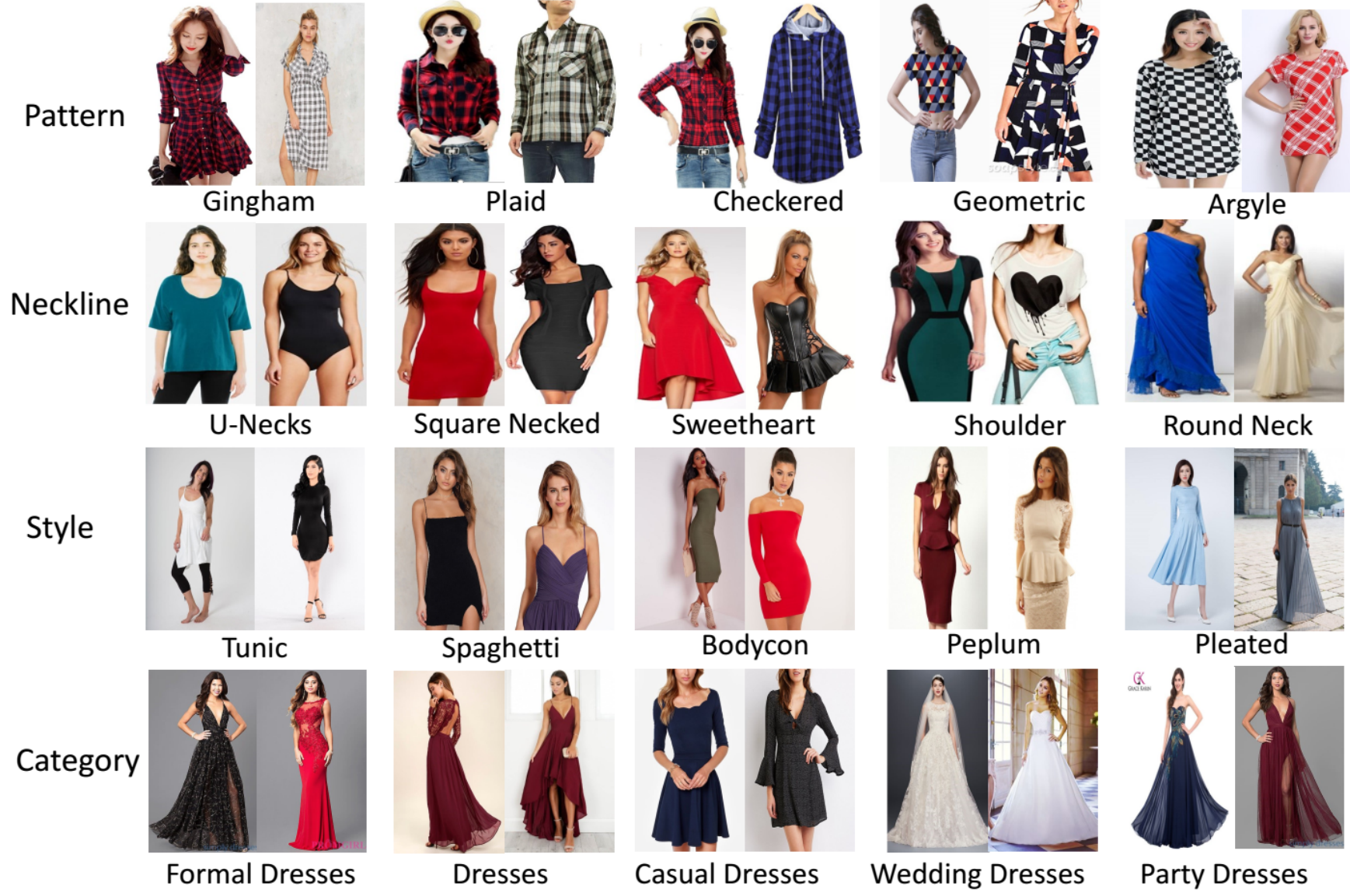}
\caption{Examples of iFashion-Attribute dataset. The attribute labels are divided into 8 groups. Here we show example images and labels from 4 attribute groups: pattern, neckline, style and category.}
\label{fig:main}
\end{figure*}

\begin{table*}[tp]
\begin{center}
\small
\begin{tabular}{p{1.35cm}<{\raggedright}|p{0.55cm}<{\centering}|p{0.6cm}<{\centering}|p{1.2cm}<{ \raggedleft}|p{2.3cm}<{ \centering}|p{8.9cm}<{\raggedright}}
\hline
\textbf{ Attribute}& \textbf{Class}& \textbf{Label}& \#\textbf{Label}& \#\textbf{Image}&\textbf{ Example} \\
\hline
Category & 105&S&913,857&913,857 - 90.2\% &Athletic Pants, Bikinis, Cargo Pants, Heels, Petticoats ...\\
Color & 21 &M&894,904&467,137 - 46.1\% &Black, Bronze, Gold, Gray, Green ... \\
Gender & 3 &M&1,012,947&935,265 - 92.3\%&Male, Female, Neutral. \\
Material & 34 &M&701,197&591,175 -58.4\%&Nylon, Organze, Patent, Plush, Rayon ... \\
Neckline & 11 &S&721,908&721,908  -71.3\%&Racerback, Shoulder Drapes, Square Necked, Turtlenecks, U-Necks ... \\
Pattern & 28 &M&325,361&311,676 - 30.8\%&Argyle, Camouflage, Checkered, Floral, Galaxy ... \\
Sleeve & 5 &S&733,501&733,501 - 72.4\%&Long Sleeved, Puff Sleeves, Short Sleeves, Sleeveless, Strapless. \\
Style & 21 &S&610,442&610,443 - 60.3\%&Asymmetric, Summer, Tunic, Vintage Retro, Wrap ... \\
\hline
\end{tabular}
\end{center}
\caption{ The detail of iFashion-Attribute groups. The table shows the number of attribute classes, label and images in each group.}
\label{tbl:iMaterialist Fashion}
\end{table*}


Secondly, many of the fashion attributes in iFashion-Attribute are fine-grained labels and may have very similar visual patterns. For example, as shown in Fig.~\ref{fig:main}, in the group of $Neckline$, identifying fine-grained visual difference on the defined fashion pattern ($Neckline$) between classes of $U$-$necks$ and $Shoulder$ is particularly challenging because the images often have large visual diversity within each class. This is much more significant than the subtle distinctions between different classes on the defined fashion pattern. The results is a large intra-class diversity which can be significantly larger than inter-class variance. This gives rise to new challenges on fine-grained image recognition compared to existing benchmarks for fine-grained recognition where images often have similar visual appearance with low intra-class diversity, such as birds in  CUB-200-2011 database~\cite{wah2011caltech} and cars in Stanford Cars database~\cite{KrauseStarkDeng}. The new dataset has 8 high-level groups of fashion attributes, each of which contains a number of fine-grained attribute-level classes. This allows it to define multiple visual patterns for fine-grained recognition, further increasing the diversity of recognition. Additionally, iFashion provides a significantly larger number of training images than existing benchmarks. These properties make the new dataset more meaningful for real-world applications by enabling the learning of CNNs with a stronger generalization capability than existing datasets.


\begin{table*}[tp]
\small
\begin{center}
\begin{tabular}{p{3.5cm}|  p{1.5cm}<{\raggedleft} | p{1.2cm}<{\centering} | p{1.5cm}<{\centering} | p{1.5cm}<{\centering} | p{1.5cm}<{\centering}| p{1.5cm}<{\centering}}
\hline
 \textbf{ Dataset} &  \textbf{\#Train}   &   \textbf{\#Classes}   &   \textbf{ Single Labeled}    &    \textbf{ Multi Labeled}    &   \textbf{Fine Grained}   &   \textbf{ Fashion Domain}  \\
 \hline
Scene15~\cite{lazebnik2006beyond}&4,485&15& \checkmark & &&\\
MIT Indoor-67~\cite{quattoni2009recognizing} & 5,360&67&   \checkmark& &&\\
ILSVRC~\cite{DengDSLL009} & 1,281,167 &1000&  \checkmark & &&\\
SUN397~\cite{xiao2010sun} & 19,850&397& \checkmark & &&\\
Places365~\cite{zhou2017places} & 8,000,000&365& \checkmark & &&\\
WebVision~\cite{li2017webvision} &2,439,574  &1000& \checkmark && &\\
\hline
NUS-WIDE\cite{nuswide} & 161,789 &81& &  \checkmark &&\\
VOC2012~\cite{everingham2015pascal}& 5,717&20  & &   \checkmark   &     &    \\
MS-COCO~\cite{lin2014microsoft}&8,787 &80& &  \checkmark &&\\
Open Images V4~\cite{kuznetsova2018open}& 1,700,000 & 600 &  & \checkmark && \\
 \hline
 CUB-200-2011~\cite{wah2011caltech}& 5,994 &200&   \checkmark && \checkmark& \\
Stanford Dogs~\cite{khosla2011novel}& 12,000 &120& \checkmark && \checkmark&\\
 Stanford Cars~\cite{KrauseStarkDeng}& 8,144 &196& \checkmark && \checkmark&\\
FGVC-Aircraft~\cite{maji2013fine}& 6,667 &100& \checkmark && \checkmark&\\
iNaturalist 2017~\cite{van2018inaturalist}&579,184 &5,089& \checkmark &&  \checkmark &\\
\hline
DCSA ~\cite{chen2012describing}&1,856 &26&& \checkmark & &\checkmark\\
ACWS~\cite{bossard2012apparel}& 145,718 &15& & \checkmark & &\checkmark\\
Clothes-1M~\cite{xiao2015learning} &  1,050,000&14&   \checkmark& &&\checkmark\\
WTBI~\cite{hadi2015buy}& 78,958 &11&& \checkmark & &\checkmark\\
 DeepFashion-C~\cite{liu2016deepfashion}& 209,222&46& \checkmark && &\checkmark\\
 DeepFashion-A~\cite{liu2016deepfashion}&209,222 &1000&& \checkmark & \checkmark &\checkmark\\
\hline
 iFashion-Attribute& 1,012,947 &228&& \checkmark &\checkmark &\checkmark\\
\hline
\end{tabular}
\end{center}
\caption{Summary of popular datasets. Each section of the table covers one type of classification task in computer vision, which are (top to bottom): single-label classification,  multi-label classification, fine-grained classification and fashion related classification. iFashion is the first known dataset which is expert-labeled at million-scale with multi-label and fine-grained attributes (see Sec.~\ref{dataset_quality}).}
\label{tbl:databases}
\end{table*}

The goal of iFashion-Attribute is to encourage research on a more complex but real-world task, by jointly considering multi-label and fine-grained image recognition with a hierarchical class structure. Our major contributions are: (i) the first known million-scale image dataset with multiple fine-grained attribute labels curated by experts; (ii) extensive experiments were conducted by using recent CNN models for multi-label and fine-grained recognition tasks, providing meaningful baseline results that facilitate future research; (iii) we demonstrate empirically that iFashion is valuable for transfer learning on other fashion related datasets and tasks.

\section{Related work}
Databases have always been a key resource for computer vision research. They provide standard benchmarks for evaluating algorithms developed for a defined task; e.g., image recognition which is a fundamental task. In this section, we review a number of recent databases created for image classification, by categorizing them into four groups: single-label, multi-label image classification, fine-grained recognition and fashion-related benchmarks.

\textbf{Single-label image classification}. There are two fundamental applications for single-label image classification: object recognition and scene classification. ImageNet~\cite{DengDSLL009} is a widely-used dataset created for large-scale object classification. It has 1000 object categories with over 1 million training images having clean human annotations. This allows it to boost the performance of recent CNNs for various computer vision tasks. However, performance on ImageNet has reached the saturation point in terms of performance with a 2.25\% Top-5 error using SE-Net~\cite{hu2018senet}. WebVision~\cite{li2017webvision} was created in 2017 by increasing data difficulty where a large amount of noisy labels is considered. Noisy labels are generated from metadata information from web images, without any human annotation or cleaning. It has 1000 object categories which are exactly the same as ImageNet, and about 2.4 million training images crawled from the Internet. Recently, Guo et al.~\cite{sheng2018} developed CurriculumNet which learns CNNs from WebVision data with a 5.2\% Top-5 error achieved. Additionally, there are a number of standard databases built for scene recognition, including Scene15~\cite{lazebnik2006beyond}, MIT-67~\cite{quattoni2009recognizing}, SUN397~\cite{xiao2010sun} and Places365~\cite{zhou2017places}. These databases are all designed for single-label image classification, and were created by increasing the number of scene categories (from 15 to 397), and the number of training images (from about 5K to 8K). Details are summarized in Table \ref{tbl:databases}. Note that it is more difficult to define a large number of scene categories than object classes, due to category ambiguity.

\textbf{Multi-label image classification}. There are common scenarios where an image may contain multiple object items. Obviously, describing such an image by just using a single object class is less informative, and multi-class descriptions should be considered. Multi-label image classification is well established in the literature with a number of databases created. For example, Everingham~\etal~\cite{everingham2015pascal} built Pascal Visual Object Classes (VOC) which is an important database for object recognition. The Pascal VOC has 20 object categories, with 5,717 training images, each of which has one or multiple labels. NUS-WIDE~\cite{nuswide}  is a web image dataset created by Chua~\etal, for evaluation of traditional image annotation and multi-label image classification.  It contains 161,789 training images with associated tags collected from Flickr. The images are manually labeled into 81 concepts, including activities, objects and scenes. Recently, Lin~\etal built the Microsoft Common Objects in Context (MS-COCO) dataset~\cite{lin2014microsoft} which is a widely-used database for object detection, (instance) segmentation, and multi-label image classification. It has 82 object categories, with 8,787 training images, each of which has been tagged with multiple labels of object categories. Our database is collected for joint multi-label and fine-grained image recognition, and is several magnitudes larger in scale.

\textbf{Fine-grained image recognition}.
Fine-grained image classification has long been studied in the computer vision community. There are a number of databases built for this task, including  CUB-200-2011~\cite{wah2011caltech}, Stanford Dogs~\cite{khosla2011novel},  Stanford Cars~\cite{KrauseStarkDeng} and FGVC-Aircraft~\cite{maji2013fine}. Training images in these datasets range from about 6K to 12K, with 100 to 200 fine-grained categories. They have common properties and the whole dataset just has a single type of object or animal (e.g., Cars, Aircrafts, Dogs, and Birds). Recently, the iNaturalist dataset was created by Hon~\etal~\cite{van2018inaturalist}. It has 859,000 images from over 5,000 different species of plants and animals, increasing both the number of training images and the number of categories considerably. Furthermore, multiple types of plants and animals are included, rather than previous versions where all images in the database have only a common type of object. Differing from these single-label databases, our iFashion-Attribute dataset is designed for both multi-label and fine-grained recognition of fashion images. Particularly, we define multiple fine-grained differences in multiple dimensions, by using various visual patterns designed in fashion applications. This allows for significant visual diversity within each fine-grained class, increasing learning complexity considerably.


\textbf{Fashion-related image recognition}.
Research efforts have recently been devoted to fashion-related tasks, such as clothes recognition~\cite{xiao2015learning,liu2016deepfashion}, fashion attribute prediction~\cite{liu2016deepfashion}, fashion retrieval~\cite{liu2016deepfashion,hadi2015buy} and clothing parsing~\cite{yamaguchi2012parsing}. Table \ref{tbl:databases} summarizes databases created for these tasks. For example, DeepFashion~\cite{liu2016deepfashion} was introduced by Liu~\etal.
It contains more than 800,000 images, where 209,222 images were used for clothes classification and attribute perdition, with 50 categories and 1,000 attributes defined. The remaining are used for other tasks, such as fashion retrieval and landmarks prediction. 
Compared with Deepfashion, the proposed iFashion-Attribute dataset has 5 times more images. In addition, our attribute labels go through several rounds of post-processing steps (see Sec.~\ref{dataset_quality}) which significantly improves label quality. In comparison, the labels in DeepFashion are crawled from the metadata on the web and does not go through any post-processing.
Xiao~\etal~\cite{xiao2015learning} built a clothes dataset (called Clothes-1M) by crawling images from several online shopping websites, which may contain a large amount of noisy labels. It has 1 million raw images (containing noisy labels generated from meta information), and 50K manually-clean and fully-annotated images from 14 clothes categories. The scale of Clothes-1M is comparable to iFashion, but it has only one label per image, the labels are noisy in the training set and the labels are coarse categores; not fine-grained. As shown in our experiments, by providing more meaningful multi-label and detailed attribute information, CNNs pre-trained on iFashion achieve better generalization, allowing it to transfer more effectively into a new task.

\section{iFashion Dataset}
We describe details of the iFashion-Attribute database in this section, including data collection, construction of the train, validation and test splits. We describe the methods for manually cleaning and labeling images in the validation and test set. Evaluation measurements are also presented.

\begin{figure}[tb]
\centering
\includegraphics[width=0.95\columnwidth]{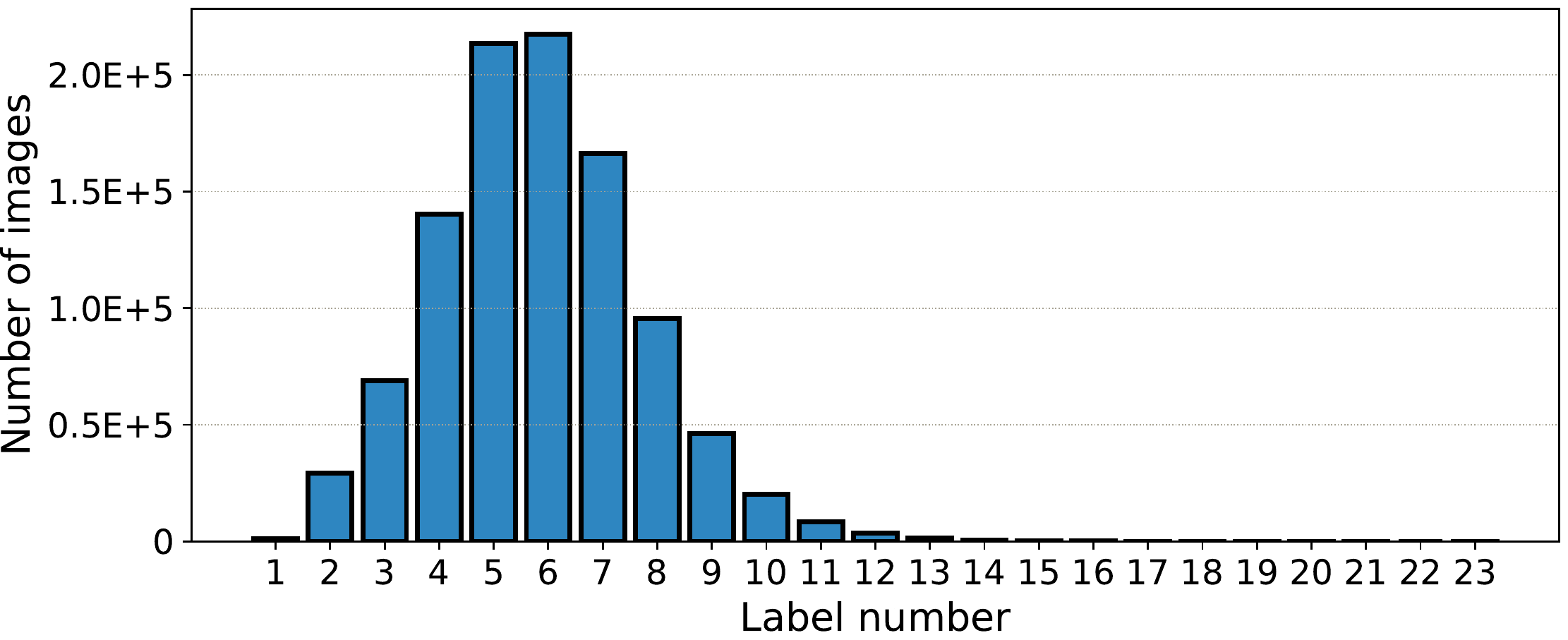}
\caption{Histogram of number of labels per image, with an average of 5.8 and 8 per image in the training and validation sets.}
\label{fig:label_num}
\end{figure}

\begin{figure}[tb]
\centering
\includegraphics[width=0.95\columnwidth]{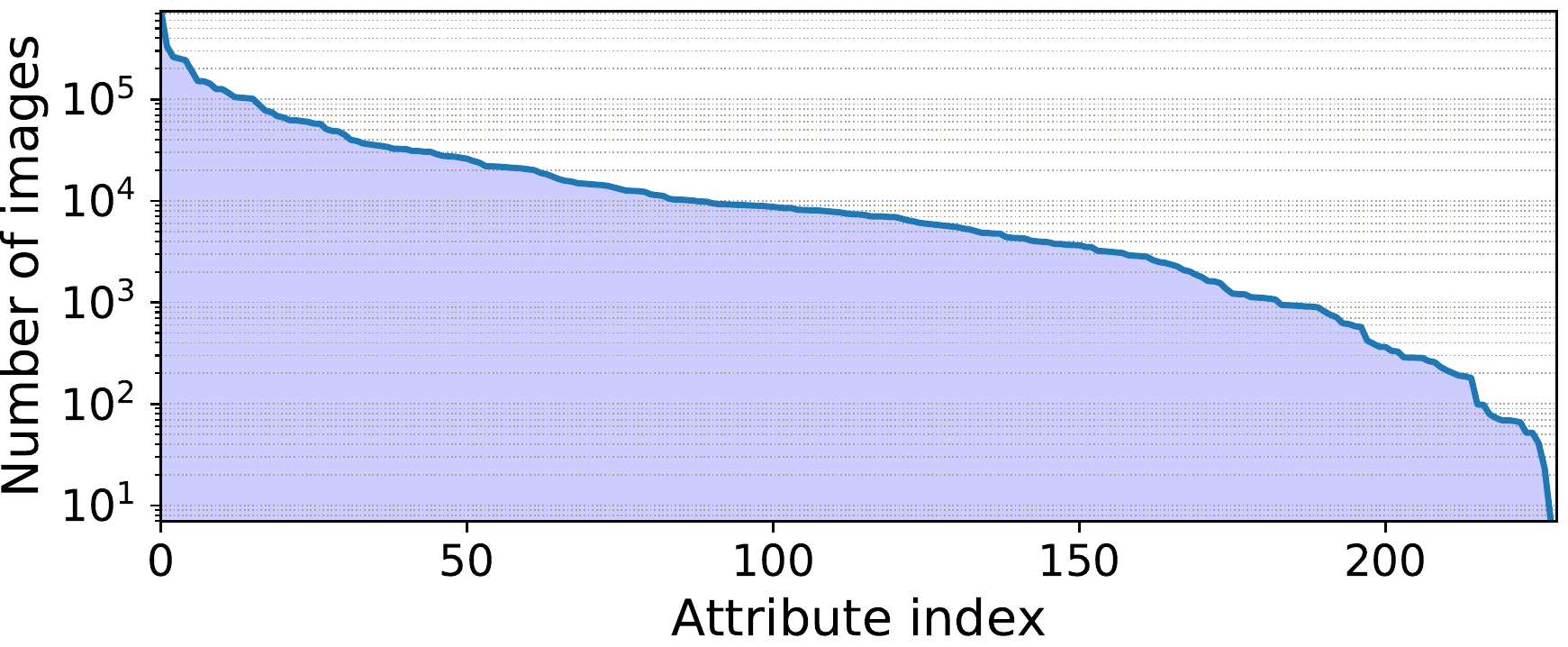}
\caption{Number of images per attribute label, demonstrating the long tail nature of the dataset.}
\label{fig:category_num}
\end{figure}


All images in the iFashion-Attribute database are provided by Wish, with 1,012,947, 9,897 and 39,706 images split into train, validation, and test sets respectively. It has 228 fine-grained fashion attribute-level classes which form 8 high-level fashion groups defined professionally from the fashion industry. Each image has multiple labels. The histogram of number of labels per image and number of images per label are shown in Fig.~\ref{fig:label_num} and Fig.~\ref{fig:category_num}.

The 8 fashion groups are presented in Table~\ref{tbl:iMaterialist Fashion}. Each group has a number of fashion classes, ranging from 3 ( ``gender'' group) to 105  ( ``category'' group).
Statistics on the number of images and labels provided for each group are listed in Table~\ref{tbl:iMaterialist Fashion}. We found that the ``Gender'' group has a label in all images, while the ``Pattern'' group just has labels in 30\% of all images in the training set. For each image, the number of labels it has, is ranged from 1 to 23, with an average of 5.8 labels for each training image. Furthermore, we also compute statistics on the number of images for each attribute-level class. There are 31 fine-grained classes, each of which has training images less than 500, while the number of classes with over 10K training images is 88.  This indicates a significant data imbalance in our database, which further increases the challenge for learning CNNs, and useful for evaluating low-shot learning algorithms.

Our database  considers large scale (million level), multiple labels (with group structure), and fine-grained recognition jointly for fashion classification, setting it apart from existing datasets which were often designed for investigating individual problems. In particular, our fine-grained classes are created structurally based on multiple groups of fashion attributes, which are professionally defined, such as ``pattern'', ``neckline'' and ``color'', as shown in Fig.~\ref{fig:main}. This allow it to have large intra-class variance which is much more significant than visual diversity between different classes (as shown in Fig.~\ref{fig:main}), making it more challenging than existing fine-grained or fashion-related benchmarks. For example, a same model may have visual similar clothes but with fine-grained distinctions on fashion pattern in two different attributes of ``Plaid'' and ``Checkered'', as shown in the ``Pattern'' group in Fig.~\ref{fig:main}. This sets a new challenge for CNNs to learn fine-grained distinctions between such structurally-defined patterns automatically from the provided data and labels.






\subsection{Dataset collection and quality improvements}\label{dataset_quality}
The fashion images are provided by Wish. Wish has over 50m+ unique images across an extensive product line. We collected 1M+ fashion images by randomly sampling across individual attribute classes. In some cases the attribute class was not large enough and a smaller number images sampled in this case. All the images were pre-tagged by humans using an organically grown taxonomy. Finally, the Wish tags were mapped to the competition taxonomy and format.
Several steps are applied to post-process the dataset and improve quality.\\

\noindent{\textbf{Deduplication:}} a deduplication pass is made to remove as many exact duplicates as possible and still maintain enough images. New images were substituted in if images were deleted. A downside of this deduplication process is that it was done for speed not complete accuracy so there are cases of near duplicate images (mirrored, cropped, color changes) that exist in the dataset. \\

\noindent{\textbf{Automatic verification:}} one downside of human based tagging is there maybe errors in the tags that were not picked up by QA. To filter out potential inconsistent tag/image pairs a second pass was made on the image tags and the product title to verify the image tags were in the right attribute class. The product title/description adds an additional level of accuracy to the tags, and adds augmented tags that were missed by human annotators.\\

\noindent{\textbf{Data checking and cleaning:}} Wish provides the original product images containing a total of 9 fashion groups, and 242 attribute-level classes. We found that some of the original attributes are ambiguous, which are difficult to be well defined and discriminated clearly by visual information, such as ``size'' group including five attributes: $‘long size, one size, short, Knee Length, Plus Size’$. We removed the whole ``size'' group with its attributes from the dataset, and reduced the number of groups from 9 to 8. Furthermore, there is a number of attribute-level classes which are defined by coarse fashion concept. For example, the attribute ``top'' included the high-level ``category'' group can include other attribute-level classes, such as  ``Polo'', ``T-shirt'', ``Vest'', and ``Jacket'' in the same group. Such high-level concept attributes include $ Top, Pant, Shirt, Sweater, Bra$ ect. It is important to ensure non-overlap between the attribute-level classes within each group. Therefore, we manually removed all such coarse-concept attributes, and all remained attribute-level classes are excluded to each other on fashion concept. This further reduces the number of attributes to 228, and the remained images are 1,012,947, 9,897 and 39,706, for train, validation, and test respectively.\\

\begin{figure}[tb]
\centering
\includegraphics[width=\columnwidth]{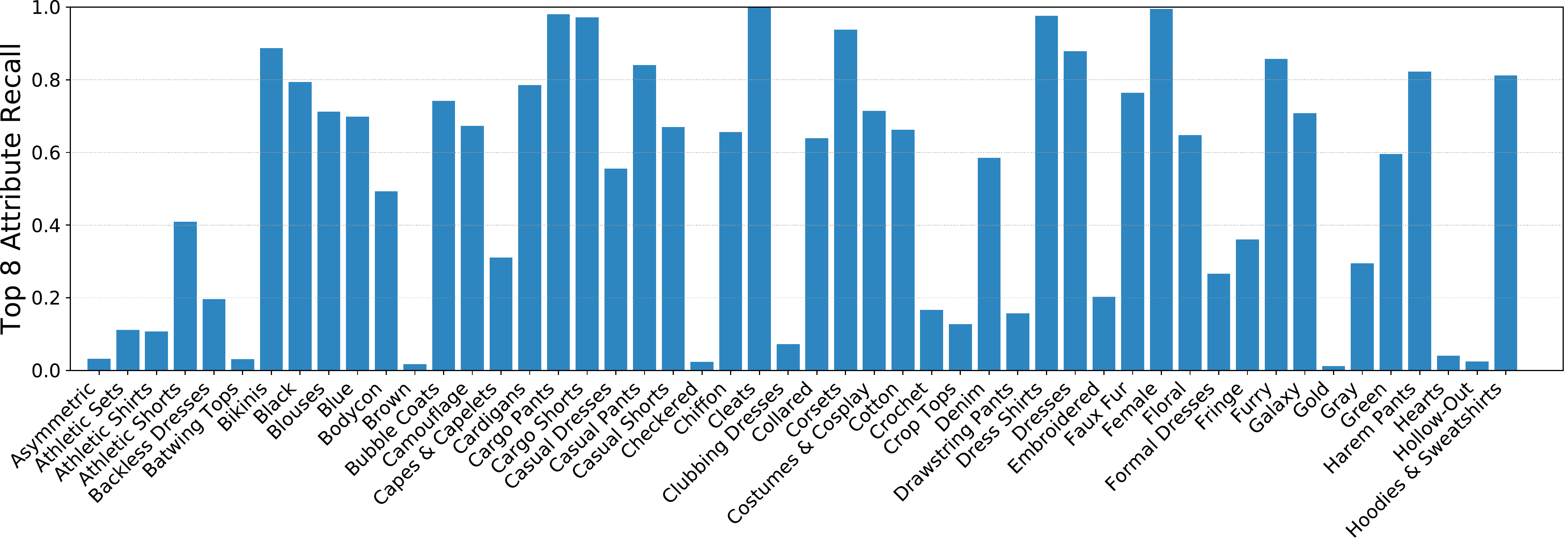}
\caption{Per-attribute recalls on 50 randomly selected attributes.}
\label{fig:per_attribute}
\end{figure}

\noindent{\textbf{Dataset statistics:}} The number of images and the number of labels provided for each group are listed in Table \ref{tbl:iMaterialist Fashion}. We found that the ``Gender'' group has a label in 92.3\% images of the training set, while the ``Pattern'' group just has labels in 30\% of all training images. For each image, the number of labels it has, is ranged from 1 to a maximum of 23, with an average of 5.8 labels over all training images. Furthermore, we also compute the number of images for each attribute-level class. There are 31 classes, each of which has training images less than 500, while the number of attributes with over 10K training images is 88.  This indicates a significant data imbalance in the database, which further increases the challenge for learning CNNs. The top-10 attributes with the largest number of training images are: $Female$, $ Long Sleeved$, $Round Neck$, $Black$, $Sleeveless$, $Male$, $White$, $V$-$Necks$, $Cotton$ and $Dresses$. In addition, recognition difficulty are changed significantly over different high-level groups or attribute-level classes, as shown in Fig.~\ref{fig:per_attribute}, where top-8 recalls for 50 randomly selected attribute-level classes are reported. The average recalls (top-1) for 8 groups are: 58.5\%, 48.3\%, 97.3\%, 52.2\%, 66.0\%, 43.1\%, 86.2\%, and 28.8\%, corresponding to the 8 groups listed in Table \ref{tbl:iMaterialist Fashion}. This further increases the challenge of our dataset.  \\

\noindent{\textbf{Validation set and test set.}} All images in validation set and test set were manually checked and annotated carefully. The two sets were created as follows. We randomly collected 10,000 and 40,000 images from \emph{ordinal} data as validation set and test set, respectively. We first processed attribute checking as did on the training set, and removed a small number of confused images with confused concepts as mentioned.
    To ensure correct labels for all images, each image was checked or re-labeled by three human labelers: (i) each image was first checked or annotated by two labelers separately; (ii) the third labeler double checked the image if its labels provided were not consistent. The last labeler made the final decision, and an image can also be discarded if it did not include a clear fashion item.
    Finally, we obtained  9,897 manually-cleaned images for the validation set, and  39,706 images for the test set.\\

\section{Experiments, Baselines and Comparisons}
We conduct experiments by using recent CNN models, and provide meaningful baseline results that facilitate future research. We analyze and discuss challenge of our database based on experimental results. Furthermore, extensive experiments are also conducted on two fashion databases (Clothes-1M~\cite{xiao2015learning} and DeepFashion~\cite{liu2016deepfashion}), for investigating the transfer learning capability of various databases. We use the CNNs trained from our dataset as pre-trained models, which demonstrate its strong generalization capability by transforming to other databases or applications.

\subsection{Baseline Results}
\textbf{ Evaluation Metrics.} We first evaluate the performance of various models for multi-label classification on our iFashion-Attribute dataset. Our measurement method was inspired by a comprehensive study from Wu~\etal~\cite{wu2016unified} on reviewing performance measurement methods for multi-label classification.  We employ a micro recall, a micro precision and  a mean-F1 score  for measuring performance on our validation and test sets. The ``micro'' means that it is an overall measure over all images. We select the top-8  classes  with the highest output score as the predicted results for each image. Specifically, the evaluation metrics are computed via precision (P) and recall (R) as follows:
 \begin{equation}
P=\frac{\sum_{i}^{C}{N_i^t}}{\sum_{i}^{C}{N_i^p}}, \quad  R=\frac{\sum_{i}^{C}{N_i^t}}{\sum_{i}^{C}{N_i^g}}, \quad F1 = \frac{2 \times P\times R}{P + R}.
\end{equation}
 Where $C$ is the number of classes, and $N_i^t$ is the number of images which are correctly predicted for the $i$-th class. $N_i^p$ is the number of predicted images, and $N_i^g$ is the number of ground truth images.  We apply the Binary Cross-entropy loss instead of softmax loss to train the CNNs,
 
\begin{equation}
L(P, Q) = -\sum_{i}^{C}p_i\log{q_i} + (1-p_i)\log{(1-q_i)}.
\end{equation}
where $C$ is the number of classes, $P$ and $Q$ denote a ground truth binary vector and the predicted probability scores.

 To facilitate future research on this task by using the iFashion-Attribute database,  we report a set of experimental results by using a number of widely-used deep network architectures, including Inception-V1~\cite{szegedy2015going}, Inception-BN~\cite{ioffe2015batch}, Inception-V3~\cite{szegedy2016rethinking}, and ResNet~\cite{HeK2016}. We apply data augmentation during training. Training images are resized as $N \times N$, where $N$ is set to 256 for Inception-V1~\cite{szegedy2015going}, Inception-BN~\cite{ioffe2015batch} and ResNet~\cite{HeK2016}, and 336 for Inception-V3~\cite{szegedy2016rethinking}.  Then, we randomly crop a $w \times h$ region at a set of fixed positions, where the cropped width $w$ and height $h$ are picked from ${N, 0.9375N, 0.875N}$.  Then these cropped regions are further resized as $M \times M$ ($M = 224$  or $299$) for training the models, where $M$ depends on the image resolution $N$, and is set as $0.9375N$. Meanwhile, we also implement a horizontal flip randomly on the cropped images. The batch size is set as 256, and the learning rate is from 0.1, decayed according to a fixed schedule determined by the dataset size.  RMSProp optimization with momentum is set to 0, and decay is set to 0.9.

\begin{table}[tp]
\small
\begin{center}
\begin{tabular}{p{1.95cm}<{\raggedright}|p{0.5cm}<{\centering}|p{0.5cm}<{\centering}|p{0.5cm}<{\centering}|p{0.5cm}<{\centering}|p{0.5cm}<{\centering}|p{0.5cm}<{\centering}}
\hline
 \multirow{2}{*}{ Method} &  \multicolumn{3}{c|}{Validation} & \multicolumn{3}{c}{Private Test} \\
\cline{2-7}
 &  R&  P&  F1&  R &  P&   F1\\
\hline
Inception-BN&59.4 &59.6&59.5 &59.0 &59.6&59.3 \\
Inception-BN$\ast$&60.0&60.2& 60.1 &59.6&60.2& 59.9 \\
Inception-v1&59.9  &60.1&60.0 &59.5  &60.1&59.8 \\
Inception-v3& 60.5&60.7&60.6 &59.9  &60.5&60.2 \\
Resnet101 & 59.7 & 59.9 & 59.8 & 59.3 & 59.9 & 59.5 \\
\hline
\end{tabular}
\end{center}
\caption{Baseline results on iFashion-Attribute with different models evaluated by precision (P), recall (R) and F1 score.}
\label{tbl:baseline}
\end{table}

 Results of various CNN models on the validation set and private test set are reported in Table  \ref{tbl:baseline} in the term of \emph{recall}, \emph{precision} and \emph{mean-F1} score.  From the Inception family, we found that a deeper network Inception-V3 outperforms the Inception-V1 and  Inception-BN on both validation and test sets.  But surprisingly, the results of Inception-V1 are slightly better than that of Inception-BN. We hypothesize that such results may due to the complexity of our database which is significantly more difficult than single-label databases for image classification where models from the Inception family were regularly applied. Therefore, training CNNs from our dataset may require more local supervised information, and Inception-V1 and Inception-V3 have multiple loss functions to enhance local supervision. This challenge has not been fully investigated in the community, and may open new research interests on multi-label image classification with hierarchical label structures.
 
In addition, we further train an ImageNet pre-trained Inception-BN$\ast$, which can improve the results from the original one.  This suggests that ImageNet pre-trained CNN features are useful to learn from our database. For data imbalance, we implemented the weighted binary cross entropy (weighted BCE) loss which was originally developed in~\cite{liu2016deepfashion} for handling this issue, and obtained a 0.7\% performance improvement on iFashion by using the Resnet architecture. Higher performance can be expected by designing more advanced approaches that jointly consider data imbalance with multi-label and fine-grained recognition problem. 

%
%

\subsection{Transfer Learning}
To investigate the generalization ability of CNN models learnt from the iFashion-Attribute dataset, we compare them by using various related databases, such as DeepFashion~\cite{liu2016deepfashion}, ImageNet~\cite{DengDSLL009} and Clothes-1M~\cite{xiao2015learning}.  Capability of the pre-trained models learnt from these databases is compared by using transfer learning, which transforms model capability to learning from a new small-scale database by fine-turning the pre-trained models. Clothes-50K~\cite{xiao2015learning} and DeepFashion~\cite{liu2016deepfashion} are used as the new databases for evaluation. Our goal here is not to evaluate the capability of various datasets, but to investigate new algorithms, CNN architecture and / or training schemes. Thus we simply apply a widely-used CNN architecture - Inception-BN~\cite{ioffe2015batch}, with regular training and fine-tuning schemes.

\subsubsection{Transfer learning on Clothes-50K~\cite{xiao2015learning}}
We evaluate the transfer ability of CNN models to Clothes-50K, which is a subset of Clothes-1M~\cite{xiao2015learning}. The Clothes-50K has 50,000 training images of 14 categories. All images were manually cleaned and annotated.  We test the trained models on the validation set, which contains 14,312 images of 14 categories.
Two groups of experiments are conducted. First, we train Inception-BN models~\cite{ioffe2015batch} individually from four different databases mentioned. Then the four models are used as pre-trained models, and we fine-tune them by using the Clothes-50K data. Second, we train Inception-BN models from the iFashion-Attribute, DeepFashion and ImageNet, and then fine-tune the pre-trained CNNs by using the Clothes-1M and Clothes-50K sequentially, by following previous approaches implemented on the Clothes-1M and Clothes-50K~\cite{xiao2015learning}. Results on the validation set of Clothes-50K are reported in Table \ref{tbl:Clothes-5K}.

\begin{table}[tp]
\begin{center}
\begin{tabular}{p{2.4cm}<{\raggedright}|p{1.5cm}<{\centering}|p{1.8cm}<{\centering}|p{1.2cm}<{\centering}}
\hline
 \textbf{ Method} & \textbf{Training Data} & \textbf{Pre-train Model} & \textbf{ Val Accuracy} \\
\hline
Inception-BN & 50K clean &ImageNet& 74.9\\
Inception-BN & 50K clean &DeepFashion& 76.4 \\
Inception-BN & 50K clean &Clothes-1M& 77.5 \\
Inception-BN & 50K clean &iFashion& 78.9\\
\hline
Inception-BN & 1M + 50K&ImageNet& 78.7\\
Inception-BN & 1M + 50K &DeepFashion& 78.3\\
Inception-BN & 1M + 50K &iFashion& \textbf{ 80.5}\\
\hline
\multicolumn{4}{c} {With new algorithms robust to noise:} \\
\hline
Xiao~\etal~\cite{xiao2015learning} & 1M + 50K & ImageNet& 78.2\\
CleanNet~\cite{lee2017cleannet} & 1M + 50K & --& 79.9\\
Patrini~\etal~\cite{patrini2017making} & 1M + 50K &--& 80.4\\
\hline
\end{tabular}
\end{center}
\caption{Transfer learning with Clothes-50K, by using pre-trained models learned from iFashion-Attribute, ImageNet, DeepFashion and Clothes-1M, with recent results by state-of-the-art algorithms.}
\label{tbl:Clothes-5K}
\end{table}

\begin{figure*}[tb]
\centering
\includegraphics[width=0.9\linewidth]{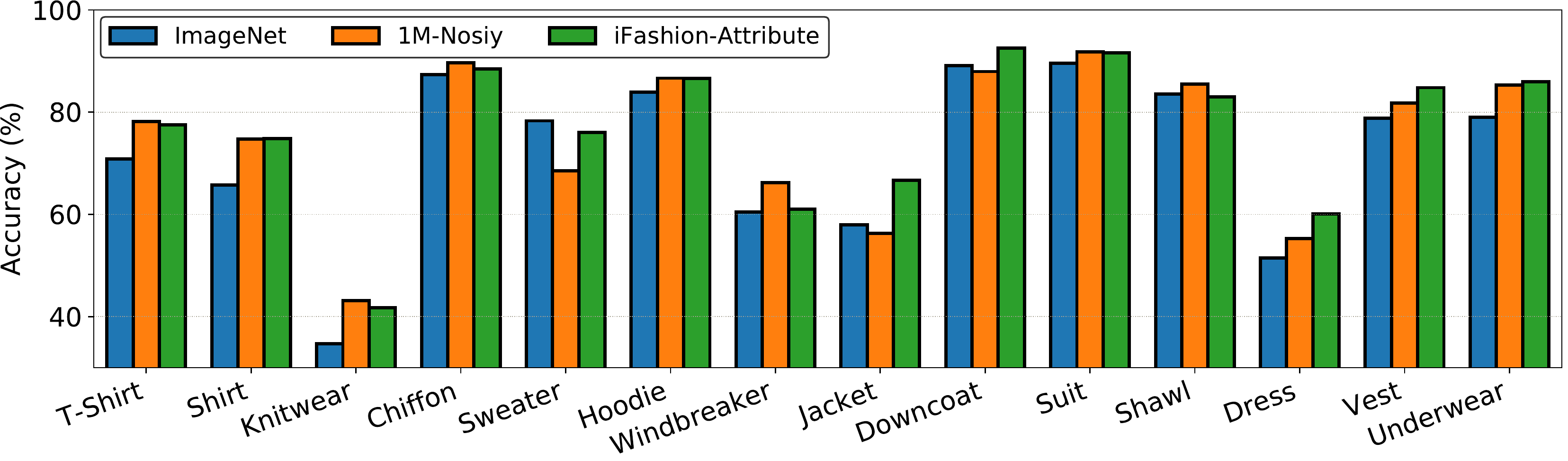}
\caption{Classification accuracy of 14 categories with three pre-training strategies (ImageNet, 1M-Noisy and iFashion-Attribute) on Clothes-1M dataset. The iFashion-Attribute pre-trained model achieves the best overall performance.}
\label{fig:1mclothes}
\end{figure*}

As shown in Table \ref{tbl:Clothes-5K}, by using Clothes-50K as training data, the pre-trained model from iFashion-Attribute obtains the best performance with an accuracy of 78.9\%. The accuracy in each class is compared in Fig.~\ref{fig:1mclothes}, where the pre-trained model from iFashion-Attribute has the highest performance in 5 classes.  It outperforms the other three pre-trained models by a large margin, particularly for ImageNet (74.9\% $\rightarrow$ 78.9\%), which just includes a number of fashion-related categories. This suggests that with a similar data scale, our database has stronger transfer capability to fashion-related tasks than the object-centralized ImageNet. Both DeepFashion and Clothes-1M are fashion-related databases. Our iFashion-Attribute has about five times the number of training images and fashion classes of DeepFashion, leading to performance improvements. Compared to Clothes-1M, we have 228 multi-label fashion classes, which is both significantly larger and more complex than 14 single-label classes, providing more meaningful supervision information for training higher performance CNNs. In addition, Clothes-1M includes a large number of noisy labels and images by crawling them raw from multiple online shopping websites. This reduces data quality, which in turn increases the difficulty of transfer learning with CNNs.

Similarly, in the second group of experiments, the pre-trained model from our database consistently outperforms those from ImageNet and DeepFashion, by using both Clothes-1M and Clothes-50K as training data. The impact of pre-trained models is decreased when the amount of training data is increased from 50K to 1M+, which naturally reduces the performance margin between different models. Furthermore, as shown in the bottom part of Table \ref{tbl:Clothes-5K}, our result of 80.5\% accuracy is even comparable or better than those of recent deep learning approaches, which were specifically designed to deal with the noisy images and labels in Clothes-1M, However with our model, empowered by our database, it just employs a simple and straightforward training method, with an off-the-shelf CNN architecture. Note that designing a new algorithm robust against noisy images and labels is beyond the scope of this work.

\subsubsection{Transfer learning on DeepFashion~\cite{liu2016deepfashion}}

\begin{table}[tp]
\small
\begin{center}
\begin{tabular}{p{3.7cm}<{\raggedright}|p{0.9cm}<{\centering}|p{0.9cm}<{\centering}}
\hline
 \textbf{ Method}   & \textbf{Top3} & \textbf{ Top5} \\
\hline
WTBI ~\cite{hadi2015buy}  &43.7& 66.3 \\
DARN~\cite{huang2015cross}  &59.5& 79.6\\
 Yang~\etal~\cite{yang2011articulated}  &75.3& 84.9\\
 FashionNet~\cite{liu2016deepfashion} &82.6& 90.2\\
\hline
 Incpetion-BN (DeepFashion)  &85.4& 91.6\\
\hline
 Incpetion-BN (Clothes-1M) &85.9& 91.9\\
 Incpetion-BN (ImageNet)  &87.3& 92.9\\
 Incpetion-BN (iFashion)  & \textbf{88.2}&  \textbf{93.3}\\
\hline
\end{tabular}
\end{center}
\caption{Transfer learning with DeepFashion, by using the pre-trained models learned from iFashion-Attribute, ImageNet and Clothes-1M, with recent results by state-of-the-art algorithms.}
\label{tbl:tranfer_deepfashion}
\end{table}

We further  evaluate the  transfer learning ability of CNN models to fashion category recognition on the DeepFahsion database~\cite{liu2016deepfashion}.  For the classification task, DeepFashion has 209,222  training images from  46 classes.  All images were manually cleaned and annotated.  We report our results on the validation set, which contains 40,000 images. Similarly, Inception-BN is used as our basic structure for experiments. 

We investigate the transfer capability of the models pre-trained on three \emph{million-level} databases: ImageNet, Clothes-1M and iFashion-Attribute. We first train three Inception-BN models individually on three databases, and then fine-tune them by using DeepFashion data. We compare these results with that of training from scratch and recent results reported in Table \ref{tbl:tranfer_deepfashion}.

As shown in  Table \ref{tbl:tranfer_deepfashion}, the results are consistent with those on Clothes-50K: (i) all pre-trained models improved the performance over that of training from scratch; (ii) iFashion-Attribute obtains the best performance on all terms, demonstrating its stronger capability for transfer learning than Clothes-1M and ImageNet; (iii) With our iFashion-Attribute for pre-training, we can achieve state-of-the-art results on the DeepFashion, by simply using  an off-the-shelf Inception-BN model. Notice that the results of the ImageNet pre-trained model are better than that of Clothes-1M pre-trained, which may be due to two reasons. First, Clothes-1M dataset includes a large amount of noisy images and labels, which may reduce its performance. A simple experiment was conducted to verify this: fine-tune the iFashion-Attribute pre-trained model on the 50K clean images from Clothes-1M, and then run the model over all training images from Clothes-1M. We obtained a correct rate of 72.7\%, by comparing the results with ground truth labels. This indicates that a large amount of images or labels from the training set are not consistent with the model prediction trained on the clean data. 
Second, many of the fashion categories presented on the DeepFashion are included in the 1000 classes of ImageNet. For example, ImageNet has 57 fashion-related categories, which include 21 overlapping categories with DeepFashion. We further investigate the impact of ImageNet pre-trained model to those overlapping and non-overlapping categories in DeepFashion, and obtained 73.8\% and 60.5\% Top-1 accuracy, respectively, compared to 67.6\% overall accuracy shown in Table \ref{tbl:tranfer_deepfashion}. 
The high-performance results on DeepFashion further confirm the promise of our database.

\section{Conclusion}
We present the iMaterialist Fashion Attributes dataset (iFashion-Attribute). It is the first known million-scale expertly curated image dataset with multi-label and fine-grained attributes. Several automated and manual processes have been taken to improve the label quality. The aforementioned characteristics of this dataset enable it to be relevant for real-world applications, particularly in fashion domain.

The introduction of iFashion-Attribute dataset allows us to compare different approaches for multi-label learning, which we provide several baselines with state of the art CNN models. Our experiments show that there is  still large room to improve in this space. We also demonstrated the value of iFashion for transfer learning. In this task, iFashion-Attribute dataset outperforms other well-known datasets for pre-trained fashion image classification models.

In future, we plan to add other annotations like bounding boxes and segmentation masks to enable localization related fashion tasks. Additionally, we plan to study few-shot learning based on the long-tail nature of the dataset.

{\small
\bibliographystyle{ieee}
\bibliography{ref}
}
\end{document}